\documentclass[lettersize,journal]{IEEEtran}  
\usepackage{graphics} 
\usepackage{epsfig} 
\usepackage{times} 
\usepackage{amsmath,amsfonts}
\usepackage{amssymb}  

\usepackage[caption=false,font=normalsize,labelfont=sf,textfont=sf]{subfig}

\usepackage{textcomp}
\usepackage{stfloats}
\usepackage{verbatim}
\usepackage{graphicx}
\usepackage{cite}
\usepackage{arydshln}
\usepackage{multirow}

\usepackage{xcolor}
\usepackage{soul}
\usepackage{fixltx2e}

\usepackage[nobiblatex]{xurl}

\usepackage{color,soul}

\usepackage{url}

\usepackage{cite}
\usepackage{caption}

\title{\LARGE \bf
A Disturbance in the Force: Force Actuation on the RAVEN II Surgical Robot with Parallel Motor-Cable Units
}

\author{Haonan Peng*, 
Dun-Tin Chiang*,
Jordan Hendricks,
Andrew Lewis,
Jared Shing,
Haokun Feng,
Yun-Hsuan Su,
Blake Hannaford, ~\IEEEmembership{Life Fellow,~IEEE} \\[1ex]
\thanks{Haonan Peng, Dun-Tin Chiang, Jordan Hendricks, Andrew Lewis, Jared Shing, Haokun Feng, and Blake Hannaford, are with the University of Washington, Seattle, WA 98195, USA.
        {\tt\small {penghn, duntic, hendrjor, alewi, shingj2, haokunf, blake}@uw.edu}}%
\thanks{Yun-Hsuan Su is with Mount Holyoke College, South Hadley, MA 01075, USA.
        {\tt\small msu@mtholyoke.edu}}%
\thanks{*These authors have equal contribution to the presented work.}
}

\begin{document}
\bstctlcite{IEEEexample:BSTcontrol}

\maketitle
\thispagestyle{empty}
\pagestyle{empty}

\begin{abstract}
Difficulty in haptic feedback for surgical robots has been a long-term problem for decades. In recent years, learning-based force estimation from robot states suggests desirable accuracy without the necessity of extra sensors. However, challenges remain in obtaining representative training data in which the robot moves in the workspace under various external forces. In this work, a parallel motor-cable system is developed. With six motor-cable units installed around the robot workspace, cables with controllable tension connected to the robot end-effector can provide the desired external force without interfering with the movement of the surgical robot. The development of the system includes motor-unit hardware, control software, sensor drivers, simulations, and more. Preliminary experiments suggest an accuracy of force actuation with errors less than 1 N. 
\end{abstract}


\section{Introduction}
Surgical robots have been used in different types of surgeries and provide accurate and reliable operations between surgeons and robots \cite{rosen2011surgical}, and improve outcomes for patients \cite{sayari2019review}. Compared to conventional minimally invasive surgeries, teleoperation of surgical robots by leader and follower control enables comfortable and intuitive control for surgeons with stereo vision. However, despite for the latest Da Vinci 5 surgical robot, many surgical robots have no haptic feedback, such as the RAVEN-II surgical robot \cite{hannaford2012raven}. Even if there are motorized leader controllers, on the follower surgical robot, all motors and encoders are mounted on the robot base instead of joints, and all joints are driven by cables, which makes it challenging to measure or estimate the external contact force on the robot end-effector. Although installing force sensors on end-effector can provide direct and accurate force sensing \cite{sosnovskaya2024tissue}, it can also introduce extra cost and complexity due to size, wiring, and sterilization. On the other hand, distal force sensing using learning based methods requires no extra sensors on the end-effector \cite{yilmaz2020neural, li2017gaussian}. However, limitations remain on obtaining representative training datasets for the learning models, in which the robots can traverse their workspace while have ground truth external force applied on the end-effector.

This paper presents a parallel motor-cable force actuation system that is able to apply force to the robot end-effector with arbitrary direction and magnitude while the robot moves in the workspace. The system mainly consists of the hardware of motor-cable units, the electronics such as sensor drivers, and the controller software. For easy implementation of the system, the localization method for motor units and a simulation is also developed. Preliminary experiments suggests that the force actuation system can apply the desired force within 1 N error in all X, Y, and Z directions. The presented system can be beneficial in collecting training data for learning-based force estimation of surgical robots.



\section{System} 
Overall, the force actuation system has 6 motor-cable units installed around the RAVEN-II surgical robot, shown in Fig. \ref{Fig_hardware}. All cables are connected to the end-effector of RAVEN-II from various directions  By controlling the motor torques and adjusting the cable tensions, the system can apply forces of arbitrary direction and magnitude to the end-effector of RAVEN-II.

\subsection{Hardware}
\begin{figure*}
\centering
\includegraphics[width=0.8\textwidth]{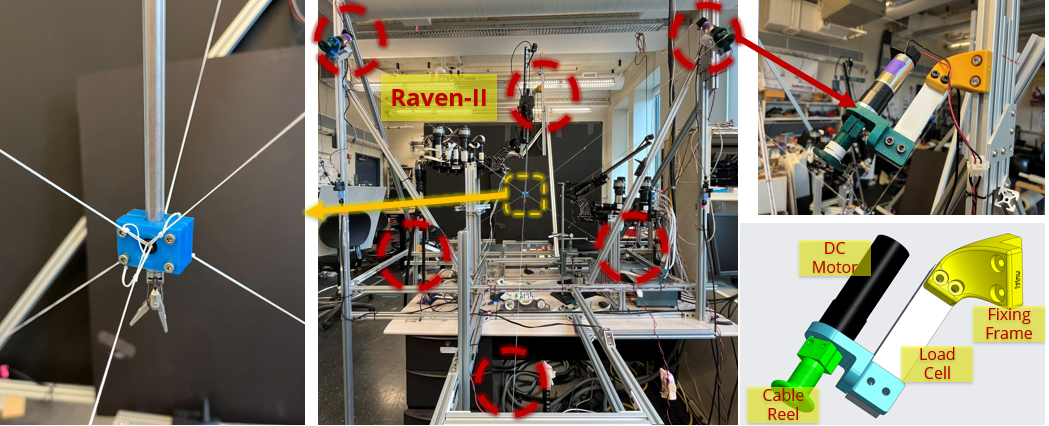}
\caption{The hardware and setup of the force actuation system. There are 6 motor-cable units (marked by red circles) installed around the robot workspace, and all cables are connected to the robot end-effector. By adjusting the tensions on cables, different forces can be applied to the end-effector.}
\label{Fig_hardware}
\end{figure*}

As shown in Fig. \ref{Fig_hardware} (right), each motor-cable units consist of the following parts: 1) a DC motor outputting torque, 2) a cable reel converting motor torque to cable tension, 3) a load cell sensing the cable tension for feedback, 4) a fixing frame that points the measurement direction of the load cells to the center of the robot workspace to make the best use of the resolution.

\subsection{Control Software}
\begin{figure*}
\centering
\includegraphics[width=0.8\textwidth]{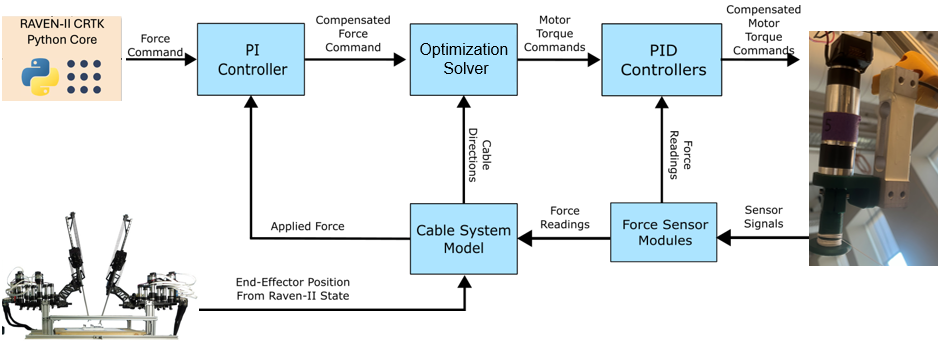}
\caption{The control workflow of the force actuation system.}
\label{Fig_control_workflow}
\end{figure*}

The control software is based on Robot Operating System (ROS) and is compatible with the RAVEN-II software controller. As shown in Fig. \ref{Fig_control_workflow}, the control loop consists of the following major procedures: 1) The system receives the force control command from the high-level controller that controls both RAVEN-II movement and force actuation. 2) Based on the motor locations and the current end-effector position from the robot state, the direction of each cable is computed. 3) Based on the force command and cable directions, 
SLSQP optimization is used to find the desired tension on each cable. A minimum cable tension is maintained all the time for the smoothness and stability of force actuation. There are 2 levels of feedback control. The higher-level feedback control is on the force command, based on the difference between the desired force and the actual applied force. The lower-level feedback control is on the cable tension, based on the difference between the desired cable tension and the actual cable tension measured by load cells. 

\subsection{Associated Electronics}
\begin{figure}
\centering
\includegraphics[width=0.4\textwidth]{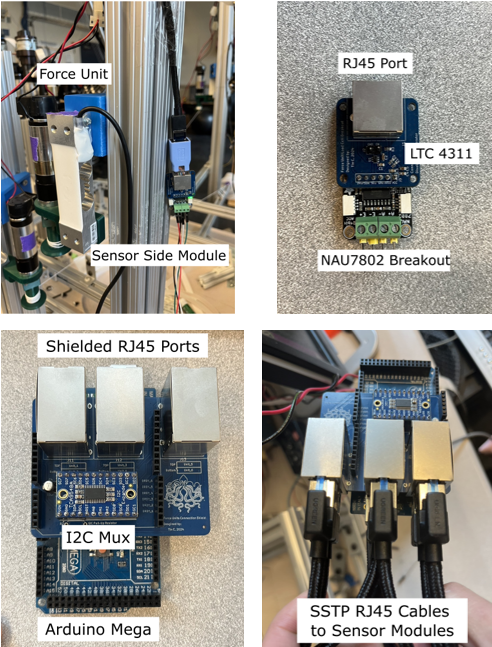}
\caption{320 Hz load cell drivers for feedback control on cable tensions.} 
\label{fig_sensor_driver}
\end{figure}

The electronics of the force actuation system are shown in Fig. \ref{fig_sensor_driver}. The load cell drivers are able to provide measurements at 320 Hz for the feedback control of the cable tensions. Currently, the motor control is achieved by the right arm of the RAVEN-II robot, with modified control software. Future work includes separating the motor control and the low-level tension control into independent motor drivers and microcontrollers.

\subsection{Simulation of the System}
\begin{figure}
\centering
\includegraphics[width=0.4\textwidth]{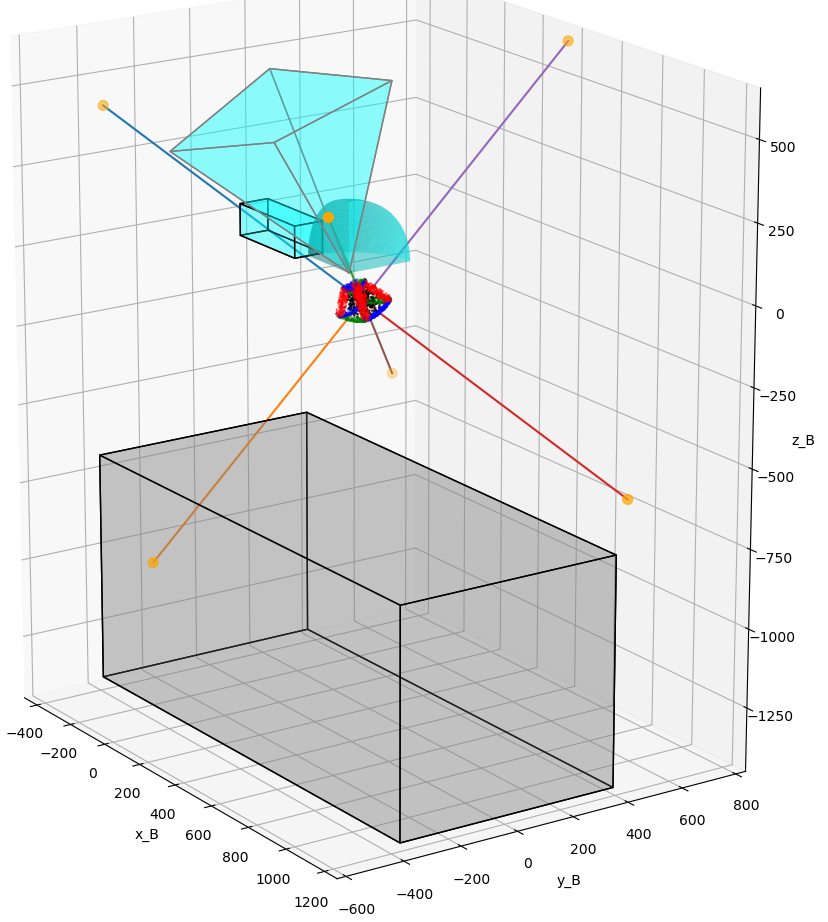}
\caption{Simulation of the force actuation system with the bounding boxes of RAVEN-II surgical robot and its workspace.} 
\label{fig_simulation}
\end{figure}

For the accuracy of the force actuation system, it is critical to find proper motor locations, so that each cable tension will not exceed the limit to provide desired forces, and cables do not interfere with the robot and other obstacles. To find desirable motor locations, a simulation of the force actuation system is developed, shown in Fig. \ref{fig_simulation}. In simulation, bounding boxes of the movement of the robot arm links are created to check the interference of cables against the robot. And the Monte-Carlo simulation of the robot end-effector workspace is also developed to analyze the maximum cable tension needed for the desired range of force actuation. Motor positions closer to the robot workspace have benefits of a smaller influence of cable elasticity, but the cable direction can have larger changes due to robot movement and results in larger cable tensions.

\subsection{Localization of Motor Units}
In order to find the cable directions in the control loop, the motor units' locations must be known in the coordinate frame of the robot. MicroScribe MX is used to measure the position of the motor units, however, the position is represented in the device's own frame. To obtain the transformation between the robot frame and the measurement device frame, with the base of the measurement device fixed, several points on the robot base with known positions in the robot frame are also measured. Next, with the same points but positions represented in the robot frame and device frame, Kabsch algorithm is used to obtain the transformation between the two frames.

\section{Preliminary Results}
\begin{figure}
\centering
\includegraphics[width=0.4\textwidth]{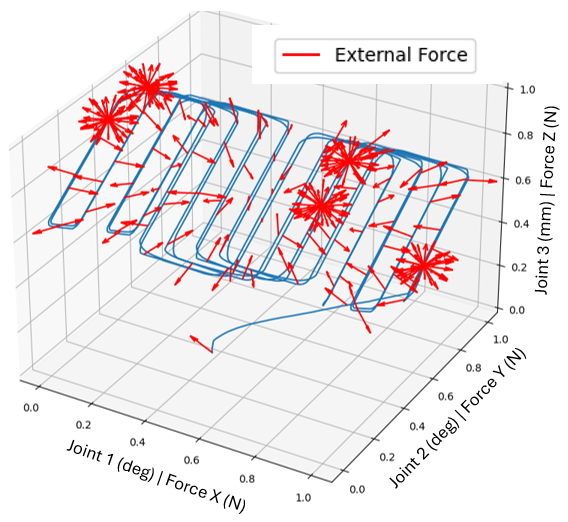}
\caption{Recorded robot trajectory with external force. The "stars" are for studying the static condition of the robot, in which the robot is not moving while the external force keeps changing. Both position and force are normalized by their ranges for better visualization.} 
\label{fig_force_traj}
\end{figure}

With the force actuation system, we recorded a dataset where the robot followed trajectories with both position and external force for the training of neural network models in the future, shown in Fig. \ref{fig_force_traj}. Statistical analysis suggested that compared to the desired external force, the actual applied force had mean absolute errors of 0.80 N, 0.79 N, and 0.44 N in X, Y, and Z direction, respectively.

\section{Conclusions and Future Work}
In this paper, a parallel motor-cable force actuation system is developed to apply desired external forces to the RAVEN-II surgical robot. The system includes motor-cable unit hardware, control software compatible with RAVEN-II software controller, sensor electronics, simulation, and motor localization. With the system, RAVEN-II trajectories with external forces are recorded. Future work includes training neural networks to estimate the external force without extra sensors. And the force actuation system can be further improved by better smoothness and stability in control, as well as separating motor drivers and MCUs to be independent from the RAVEN-II system.

\bibliographystyle{IEEEtran}
\bibliography{IEEEabrv,IEEEexample}

@IEEEtranBSTCTL{IEEEexample:BSTcontrol,
  CTLdash_repeated_names    = "no"
}

@article{sayari2019review,
  title={Review of robotic-assisted surgery: what the future looks like through a spine oncology lens},
  author={Sayari, Arash J and Pardo, Coralie and Basques, Bryce A and Colman, Matthew W},
  journal={Annals of translational medicine},
  volume={7},
  number={10},
  year={2019},
  publisher={AME Publications}
}

@article{hannaford2012raven,
  title={Raven-II: an open platform for surgical robotics research},
  author={Hannaford, Blake and Rosen, Jacob and Friedman, Diana W and King, Hawkeye and Roan, Phillip and Cheng, Lei and Glozman, Daniel and Ma, Ji and Kosari, Sina Nia and White, Lee},
  journal={IEEE Transactions on Biomedical Engineering},
  volume={60},
  number={4},
  pages={954--959},
  year={2012},
  publisher={IEEE}
}

@article{rosen2011surgical,
  title={Surgical robotics},
  author={Rosen, Jacob and Hannaford, Blake and Satava, Richard M},
  journal={Medical Devices: Surgical and Image Guided Technologies},
  volume={1},
  pages={301--306},
  year={2011}
}

@phdthesis{sosnovskaya2024tissue,
  title={Tissue Characterization With Surgical Smart Grasper and Hybrid CNN--GRU Model},
  author={Sosnovskaya, Yana},
  year={2024},
  school={University of Washington}
}

@inproceedings{yilmaz2020neural,
  title={Neural network based inverse dynamics identification and external force estimation on the da Vinci Research Kit},
  author={Yilmaz, Nural and Wu, Jie Ying and Kazanzides, Peter and Tumerdem, Ugur},
  booktitle={2020 IEEE International Conference on Robotics and Automation (ICRA)},
  pages={1387--1393},
  year={2020},
  organization={IEEE}
}

@article{li2017gaussian,
  title={Gaussian process regression for sensorless grip force estimation of cable-driven elongated surgical instruments},
  author={Li, Yangming and Hannaford, Blake},
  journal={IEEE robotics and automation letters},
  volume={2},
  number={3},
  pages={1312--1319},
  year={2017},
  publisher={IEEE}
}
\end{document}